\documentclass{article}
\usepackage{ijcai24}

\usepackage{times}
\usepackage{soul}
\usepackage{url}
\usepackage{xspace}
\usepackage{adjustbox}
\usepackage{multirow}
\usepackage{natbib}
\usepackage[hidelinks]{hyperref}
\usepackage[utf8]{inputenc}
\usepackage[small]{caption}
\usepackage{graphicx}
\usepackage{amsmath}
\usepackage{amsthm}
\usepackage{booktabs}
\usepackage{algorithm}
\usepackage{algorithmic}
\usepackage{amsmath,amssymb,amsfonts}
\usepackage{mathtools}

\title{Large language models can be zero-shot anomaly detectors for time series?}

\newcommand{\gpt}{\textsc{GPT\xspace}}
\newcommand{\llama}{\textsc{Llama-2}\xspace}
\newcommand{\mistral}{\textsc{Mistral}\xspace}
\newcommand{\llmad}{\textsc{SigLLM}\xspace}

\newcommand{\prompt}{\textsc{Prompter}\xspace}
\newcommand{\detect}{\textsc{Detector}\xspace}

\author{
Sarah Alnegheimish$^1$
\and
Linh Nguyen$^1$\and
Laure Berti-Equille$^2$\And
Kalyan Veeramachaneni$^1$\\
\affiliations
$^1$MIT\\
$^2$IRD ESPACE-DEV\\
\emails
\{smish, linhnk, kalyanv\}@mit.edu,
laure.berti@ird.fr,
}

\begin{document}

\maketitle

\begin{abstract}
Recent studies have shown the ability of large language models to perform a variety of tasks, including time series forecasting. The flexible nature of these models allows them to be used for many applications. In this paper, we present a novel study of large language models used for the challenging task of time series anomaly detection. This problem entails two aspects novel for LLMs: the need for the model to identify part of the input sequence (or multiple parts) as anomalous; and the need for it to work with time series data rather than the traditional text input. We introduce \llmad, a framework for time series anomaly detection using large language models. Our framework includes a time-series-to-text conversion module, as well as end-to-end pipelines that prompt language models to perform time series anomaly detection. We investigate two paradigms for testing the abilities of large language models to perform the detection task. First, we present a prompt-based detection method that directly asks a language model to indicate which elements of the input are anomalies. Second, we leverage the forecasting capability of a large language model to guide the anomaly detection process. We evaluated our framework on 11 datasets spanning various sources and 10 pipelines. We show that the forecasting method significantly outperformed the prompting method in all 11 datasets with respect to the F1 score. Moreover, while large language models are capable of finding anomalies, state-of-the-art deep learning models are still superior in performance, achieving results 30\% better than large language models.
\end{abstract}

\vspace{-0.2cm}

\section{Introduction}

Large language models (LLMs) have demonstrated an outstanding ability to learn natural language tasks implicitly, whether performing reading comprehension, text summarization, translation, or related tasks.~\cite{radford2019multitasklearners, brown2020gpt3, sanh2022t0, chowdhery2023palm, wei2022emergent}.
Moreover, LLMs have shown tremendous promise for formal language generation, including code generation and synthesis~\cite{xu2022code1, austin2021code2, chen2021code3}, and in production beyond textual output, such as generating images and videos from natural language descriptions~\cite{saharia2022image1, koh2023image2}. 
Testing these models on new tasks and data modalities allows us to push the boundaries of LLMs and discover their value.
In this paper, we present a thorough study of LLMs used for the challenging task of anomaly detection from time series data, asking the question \textit{Can LLMs become anomaly detectors for time series data?} Here, LLMs are exposed to a new data type -- time series -- and are tasked with a detection task -- different from the classification tasks in which they are known to excel at~\cite{howard2018universaltextclassifiers}.

\begin{figure}[t]
    \centering
    \vspace{-0.2cm}
    \makebox[\textwidth][l]{%
    \hspace{-0.5cm}
    {\includegraphics[height=3.8cm]{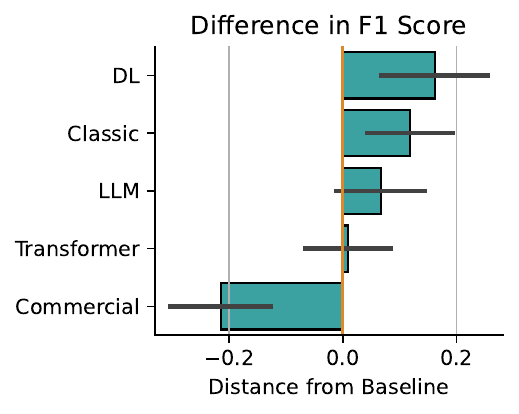}}
    \hspace{-0.5cm}
    {\includegraphics[height=3.8cm]{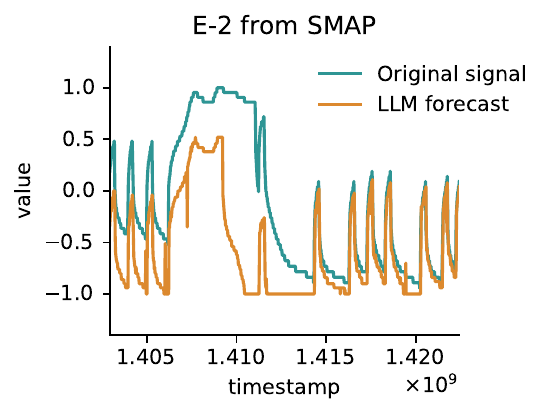}}}
    \vspace{-0.8cm}
    \caption{(left) F1 Score performances of different model types, compared to a moving average baseline. Each category represents a collection of models that fall under that group. For \textit{classic} models, we consider ARIMA and Matrix Profiling; for \textit{Deep Learning (DL)}, we utilize AER and LSTM DT; for \textit{transformer} anomaly detection models, we look at Anomaly Transformer; lastly, for the \textit{commercial} category, we compare to MS Azure. (right) Illustration of \mistral forecasts on E-2 signal from the SMAP dataset. The deviation between the signals can help identify anomalous regions.}
    \label{fig:preview}%
    \vspace{-0.5cm}
\end{figure}

Recently, \citet{gruver2024llmtime} posited that large language models have an inherent auto-regressive feature which allows them to be effective forecasters. In the study, the authors fed a string representation of a time series sequence to a pretrained LLM. The LLM then generated the next expected values, treating time series forecasting as a next-token prediction task. A follow-up question ari QWERT]\'ses: Does LLMs' auto-regressive nature allow them to take on more complex tasks, such as anomaly detection? State-of-the-art time series anomaly detection models using deep learning typically include a forecasting model as one of the steps in their process~\cite{hundman2018lstmdt}.

Time series anomaly detection is a regular part of day-to-day industry operations. Identifying unusual patterns can be a cumbersome and difficult task, especially when massive amounts of signal must be analyzed. If LLMs are genuine anomaly detectors, and can be employed directly in \textit{zero-shot} (without any additional training), they could serve as off-the-shelf anomaly detectors for users, lifting a considerable amount of this burden. Considering that training deep learning models is time-consuming, skipping this phase could make anomaly detection more efficent overall.  

In this paper, we present \llmad, a framework for using LLMs to detect anomalies in time series data, with current interaction support for models hosted by OpenAI~\footnote{\url{https://platform.openai.com/docs/models}} and HuggingFace~\footnote{\url{https://huggingface.co/models}}. Our framework includes a signal-to-text representation component to convert time series data into LLM-ready input. Moreover, we present two distinct approaches to investigating our main question. First, \prompt is a simple and direct prompting method, which elicits LLMs to identify the parts of a sequence it thinks are anomalous. Second, \detect leverages LLMs' ability to forecast time series to find anomalies, by using the residual between the original signal and the forecasted one.

Our findings, captured in Figure~\ref{fig:preview}(left), show that LLMs improve on a simple moving average baseline. Moreover, they outperform transformer-based models such as Anomaly Transformer~\cite{xu2022anomalytransformer}. However, there is still a gap between classic and deep learning approaches and LLMs.
Furthermore, between our two approaches, \detect is superior to \prompt, with an improvement of 135\% in F1 Score, as the latter suffers from false positives. We highlight the potential of the \detect approach in Figure~\ref{fig:preview}(right), which showcases an example of an LLM forecast. We can clearly see that the LLM forecast is substantially different from the original signal; this difference is attributed to the presence of anomalies.

We summarize our contributions as follows:
\begin{itemize}
    \item \textbf{Propose a new application for LLMs--anomaly detection--and study their efficacy and efficiency for this task.} We formalize a new task to present to LLMs; namely, time series anomaly detection in \textit{zero-shot}.
    \item \textbf{Present the \llmad framework with a time-series-to-text representation module and two novel methodologies for solving this task.} We present \llmad with a module to convert time series data into language-model-ready input through a series of reversible transformations. Moreover, we propose two distinct approaches for solving the problem: the \prompt pipeline and the \detect pipeline. As of this writing, our framework integrates propriety models such as \gpt{} by OpenAI and open models provided on the HuggingFace \texttt{transformers} package.
    \item \textbf{Provide a comprehensive and thorough evaluation of LLM performance on this task.} We conduct our experiments on two prominent LLM models -- \gpt{-3.5-turbo} and \mistral{-7B-Instruct-v0.2} -- and 11 datasets. We show that LLMs are able to find anomalies with an average F1 score of 0.525. Moreover, we compare \llmad methods to 10 other existing methods including state-of-the-art models such as AER~\cite{wong2022aer}.
    \item \textbf{Publish an open source software.} Our code and datasets are publicly available on github: 
    \url{https://github.com/sintel-dev/sigllm}.
\end{itemize}
\section{Background and Related Work}

\begin{figure}[t]
    \centering
    \includegraphics[width=\linewidth]{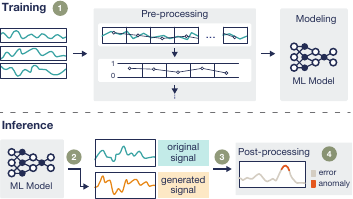}
    \vspace{-0.5cm}
    \caption{General principle of how machine learning models find anomalies in an unsupervised setting. Step 1: Apply a sequence of preprocessing operations and train a machine learning model to learn the pattern of the data. This is the most time-consuming step; Step 2: Use the trained model to generate another time series; Step 3: Quantify the error between what the model expects and the original time series value; Step 4: Use this discrepancy to extract anomalies.}
    \label{fig:how-ml-works}
\end{figure}

\noindent\textbf{Anomaly Detection Pipelines.}
Unsupervised machine learning-based anomaly detection pipelines generally follow the same sequence of steps, which roughly consist of pre-processing, modeling, and post-processing abstractions as presented in Figure~\ref{fig:how-ml-works}~\cite{alnegheimish2022sintel}. Pre-processing operations include scaling the time series into a specific range, while post-processing includes computing discrepancies between two sequences. A wide variety of models can be trained to learn the features of input data, including Long-Short-Term-Memory models (LSTMs)~\cite{hundman2018lstmdt}, AutoEncoders (AE)~\cite{malhotra2016lstmae}, Variational AutoEncoders (VAE)~\cite{park2018vae}, Generative Adversarial Networks (GANs)~\cite{geiger2020tadgan}, and Transformers~\cite{xu2022anomalytransformer, tuli2022tranad}. These models perform well on existing benchmarks such as~\cite{alnegheimish2024orionbench}, surpassing the performance of statistical approaches such as ARIMA~\cite{box1970arima, pena2013arima}.

\noindent\textbf{Transformers for Time Series.}
Transformers can be used directly to reconstruct time series~\cite{tuli2022tranad}. Moreover, the attention mechanism can be leveraged to find anomalous sequences~\cite{xu2022anomalytransformer}.
Recently, more work has emerged adopting transformer-based models for forecasting purposes by pretraining large transformer models on a large corpus of time series data. \textsc{ForecastPFN}~\cite{dooley2023forecastpfn} pre-trains a basic encoder-decoder transformer with one multi-head attention layer and two feedforward layers on a synthetically generated time series dataset. Similarly, \textsc{TimeGPT} was pretrained on a large collection of publicly available time series datasets. \textsc{Lag-Llama}~\cite{rasul2023lagllama} is a decoder-only \llama model pretrained on a large corpus of real time series data from diverse domains.
Moreover, \textsc{Chronos}~\cite{ansari2024chronos} adopts a \textsc{T5} architecture, and parses time series data into text to pretrain their model.
Most of these models were developed with the objective of creating a time series foundation model for time series forecasting.

\noindent\textbf{LLMs for Time Series.}
The past several months have seen considerable efforts toward LLM utilization for time series data. Given the parallels between predicting the next word in a sentence and predicting the next value in a time series, most of these efforts have focused on time series forecasting.
One notable effort is \textsc{LLMTime} where~\citet{gruver2024llmtime} employ \gpt~\cite{brown2020gpt3}, and \llama~\cite{touvron2023llama2} models to forecast time series data. \textsc{PromptCast}~\cite{xue2023promptcast} is a related work that translates a forecasting problem into a prompt, transforming forecasting into a question-answering task.

\begin{figure*}[t]
    \centering
    \includegraphics[width=0.9\textwidth]{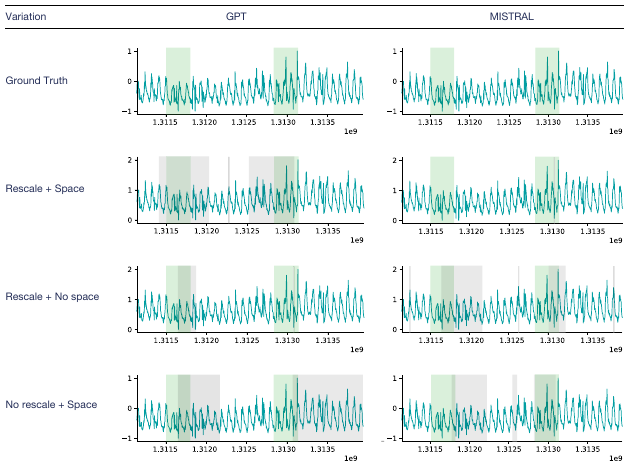}
    \caption{Visualizing the output of large language models (\gpt{} and \mistral) under different variations of the transformation process. Each row depicts the \texttt{exchange-2\_cpm\_results} signal from the AdEx dataset, where the x-axis shows the timestamp and the y-axis is the signal value. The first row indicates the ground truth anomalies present in the time series (highlighted in green). The remaining rows indicate whether scaling and inserting space between digits has occurred during the conversion from signal to text. The gray intervals highlight the anomalies detected under these conditions; thus, we would like to maximize the overlap between the green and gray intervals. Overall we find that ``scaling + space'' is the configuration that yields a better output for \gpt; and ``scaling + no space'' is better for \mistral.}
    \label{tab:justifcation_for_preprocessing}
\end{figure*}

\noindent\textbf{Our Work.} In this paper, we work strictly with LLMs that have been pre-trained on text, particularly a proprietary model using \gpt{-3.5}~\cite{brown2020gpt3} and an open source model using \mistral~\cite{jiang2023mistral}. Our main objective is to determine whether LLMs have the ability to directly uncover anomalies in time series data. Referring back to Figure~\ref{fig:how-ml-works}, our methodology focuses on Step 2 onwards -- primarily the inference phase.
To our knowledge, there is no other work that utilizes large language models as zero-shot anomaly detectors for time series data. We explore two avenues for accomplishing this task: (a) Through the paradigm of prompt engineering; (b) By leveraging LLMs' ability to forecast time series in \textit{zero-shot} without any additional data or fine-tuning.

\section{Time Series Representation}
\label{sec:timeseries-rep}
Time series data can take many different forms. In this paper, we define a univariate time series as $\mathbf{X} = (x_1, x_2, \dots, x_T)$, where $x_t \in \mathbb{Z}_{\geq 0}$ is the value at time step $t$, and $T$ is the length of the series. To make a time series LLM-ready, we transform the univariate time series $\mathbf{X}$ into a sequence of values that is tokenized. We follow a sequence of reversible steps, beginning with scaling, quantization, and processing the time series into segments using rolling windows, and ending with tokenizing each window. We detail these steps below.

\noindent\textbf{Scaling.}
Time series data includes values of varying numerical magnitudes, and may include both positive and negative values. To standardize the representation and optimize computational efficiency, we subtract the minimum value from the time series $x_{s_t} = x_t - min(x_1, x_2, \dots, x_T)$, resulting in a new time series $\mathbf{X}_s = (x_{s_1}, x_{s_2}, \dots, x_{s_T})$, where $x_{s_t} \in \mathbb{R}_{\geq 0}$. In other words, we introduced a mapping function: $\mathcal{E}: \mathbb{R} \rightarrow \mathbb{R}_{\geq 0}$. This eliminates the need to handle negative values separately. 

\begin{figure*}[!t]
    \centering
    \includegraphics[width=0.95\textwidth]{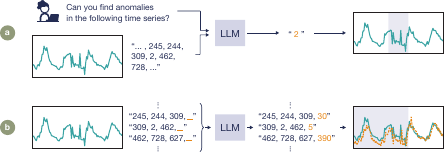}
    \caption{Anomaly detection methods in the \llmad framework. (a) \prompt: a prompt engineering approach to elicit large language models to identify parts of the input which are anomalies. (b) \detect: a forecasting approach to use large language models as forecasting methods. \detect then finds discrepencies between the original and forecasted signal, which indicate the presence of anomalies.}
    \label{fig:methods}
\end{figure*}

Other scaling methods, such as min-max scaling, can be utilized to achieve the same goal. However, reducing the set of possible values to a smaller range (e.g. [0, 1]), may cause a loss of information in the quantization step. On the other hand, increasing the range will mean there are more digits to tokenize. With our approach, we simply shift the range of the signal values, which allows us to reduce the number of individual digits that need to be tokenized while maintaining the original gaps between pairs of entries. Moreover, by projecting the values into a non-negative range, we eliminate the need for sign indicator ``-/+'' and save an additional token. 


\noindent\textbf{Quantization.}
Unlike the finite set of vocabulary words used to train LLMs (32k vocab tokens for \mistral{})~\footnote{The exact vocabulary size for \gpt{-3.5-turbo} has not been released by OpenAI.}, the set of scaled time series values $x_{s_t}$ is infinite, and cannot be processed by language models. Therefore, time series that are to be used with LLMs are generally quantized \cite{ansari2024chronos, gruver2024llmtime}. We use the rounding method, as proposed in in~\cite{gruver2024llmtime}. Because in some cases the number of decimal digits are redundant given a fixed precision, we round each value up to a predetermined number of decimals, and subsequently scale to an integer format to avoid wasting tokens on the decimal point. Hence, the input time series becomes $\mathbf{X}_q = (x_{q_1}, x_{q_2}, \dots, x_{q_T})$, where $x_{q_t} \in \mathbb{Z}_{\geq 0}$. Below is an example of this operation :
\begin{align*}
    0.2437, 0.3087, 0.002, 0.462 \rightarrow \text{``244,309,2,462''}
\end{align*}

Overall, we use 2 mapping functions: the scaling function noted $\mathcal{E}: \mathbb{R} \rightarrow \mathbb{R}_{\geq 0}$ and the quantization function noted $\mathcal{Q}: \mathbb{R}_{\geq 0} \rightarrow \mathbb{Z}_{\geq 0}$. Because both mapping functions are reversible up to a certain number of precision digits, we can always reconstruct the input time series: $\mathcal{E}^{-1} \left(\mathcal{Q}^{-1}\left(x_{q_t}\right)\right) \approx x_t$

\noindent\textbf{Rolling windows.}
Because there is an upper limit on the context length input to LLMs (e.g., \mistral has an upper limit of 32k tokens and \gpt{-3.5-turbo} has a limit of 16k tokens), and there are constraints on GPU memory, a rolling windows technique is employed to manage input data that exceeds these thresholds. This method involves segmenting each time series into rolling windows characterized by predetermined lengths and step sizes; i.e., a processed time series $\mathbf{X}_q$ is segmented and turned into a set $\{\left(x^i_{q_{1\dots w}}\right)\}_{i=1}^N$, where $w$ is the window size and $N$ is the number of windows. 
For a cleaner notation, we refer to the set as $\{\left(x^i_{1\dots w}\right)\}_{i=1}^N$. We drop $q$ in the notation from this point on, as all the input is now quantized.

\noindent\textbf{Tokenization.}
Different tokenization schemes vary in how they treat numerical values. Several open-source LLMs, such as \llama \cite{touvron2023llama2} and \mistral~\cite{jiang2023mistral}, utilize the SentencePiece Byte-Pair Encoding tokenizer \cite{touvron2023llama2}, which segments numbers into individual digits. However, the \gpt{} tokenizer tends to segment numbers into chunks that may not correspond directly with the individual digits~\cite{liu2023goat}. For instance, the number 234595678 is segmented into chunks [234, 595, 678] and assigned token IDs [11727, 22754, 17458]. Empirical evidence suggests that this segmentation impedes the LLM's ability to learn patterns in time series data \cite{gruver2024llmtime}. To make sure \gpt{} tokenizes each digit separately, we adopt the approach introduced by~\citet{gruver2024llmtime}, which inserts spaces between the digits in a number. 

Continuing with the running example:
\begin{align*}
    \text{``244,309,2,462''} \rightarrow 
    \text{``2 4 4 , 3 0 9 , 2 , 4 6 2''}
\end{align*}

Where each digit is now encoded separately. \\
Figure \ref{tab:justifcation_for_preprocessing} shows how different preprocessing steps affect the output of the model. Overall, we find that scaling reduces the number of tokenized digits, and yields better results than not scaling. Moreover, \gpt{} performs better with added space between digits, while \mistral does not. These results accord with the forecasting representation presented in~\cite{gruver2024llmtime}.

\section{\llmad: Detecting Anomalies in Signals using Large Language Models}


Given a univariate time series $\mathbf{X} = (x_1, x_2, \dots, x_T)$, and assuming there exists a set of anomalies of varied length $\mathbf{A} = \{(t_s, t_e)^i\:|\: 1 \leq t_s < t_e \leq T\}_{i=1}^m$ that is unknown \textit{a priori}, our goal is to find a set of $m$ anomalous time segments, where $t_s$ and $t_e$ represent the start and end time points of an anomalous interval. 
We introduce two fundamentally different methods that can be used for anomaly detection with LLMs: \prompt and \detect, as visualized in Figure~\ref{fig:methods}.

\begin{table*}[ht]
    \caption{Examples of prompts used in \prompt with their respective observed output. $\{x_{1..w}\}$ is a placeholder of the actual signal values in the given window.}
    \centering
    \begin{adjustbox}{max width=\textwidth,center}
    \begin{tabular}{lll}
    \toprule
    Trial & Prompt      &   Observed Output \\
    \midrule
    1 & $\{x_{1..w}\}$. Find the anomalies of the time series above.                &  (1) generating code with generic stack overflow code for anomaly detection \\
      &                                                                             & in python with numpy's \texttt{convolve}~\footnotemark{} or sklearn's \texttt{IsolationForest}~\footnotemark{}. \\
      &                                                                             &  (2) could not find anomalies \\
      &                                                                             &  (3) produced a vague answer about common approaches to finding anomalies \\
    \midrule
    2 & Find the range of indices that are anomalous in this series $\{x_{1..w}\}$ or   &  (1) producing a list of indices \\
      & Given this series $\{x_{1..w}\}$. Find the range of indices that are anomalous  &  (2) generating code similar to trial \#1 \\
      &                                                                             &  (3) could not find anomalies \\
      &                                                                             &  (4) produced a vague answer about common approaches to finding anomalies \\
      &                                                                             &  (5) asked `do you have any criteria or specific method in mind'\\
      &                                                                             &  (6) confirmed that anomalies are values deviating significantly from the mean.\\
      &                                                                             &       After confirming,  the model digressed from the topic \\
    \midrule
    3 & Find the anomalous indices in this series $\{x_{t_s-100..t_e+100}\}$.           &  (1) producing a list of indices \\
      & where $t_s$ and $t_e$ is the index of where the anomalies starts and ends, respectively.                        &  (2) could not find anomalies \\
    \midrule
    4 & The anomaly indices in timeseries\_1 = $\{x_{1..w}\}_1$ \text{ is: } $\{t_{1..k}\}_1$   &  (1) producing a list of indices \\
      & The anomaly indices in timeseries\_2 = $\{x_{1..w}\}_2$ \text{ is: } $\{t_{1..k}\}_2$   &  (2) claimed anomalies of timeseries\_3 had been given \\
      & The anomaly indices in timeseries\_3 = $\{x_{1..w}\}_3$ \text{ is: }                    &  (3) could not find anomalies\\
      &                                                                                     &  (4) outputted `Whoa, that's quite a lengthy time series!\\
      &                                                                                     &      What can I help you with regarding this data'\\
    \midrule
    5 & You are a helpful assitant that performs time series anomaly detection.                 &  \gpt-3.5-turbo:\\
      &                                                                                         &  (1) producing a list of indices \\
      & The user will provide a sequence and you will give a list of indices that are           &  (2)          occasionally, words like `Index:' were included\\
      & anomalous in the sequence. The sequence is represented by decimal strings               &  (3) sometimes, the output indices exceeded sequence length\\
      & separated by commas. Please give a list of indices that are anomalous in the            &     \mistral: \\
      & following sequence without producing any additional text. Do not say anything           & (1) produced a list of \textbf{values}.\\
      & like `the anomalous indices in the sequence are', just return the numbers. \\
      & Sequence: $\{x_{1..w}\}$\\
    \bottomrule
    \end{tabular}
    \end{adjustbox}
    \label{tab:prompt_examples}
\end{table*}

\subsection{\prompt: Finding Anomalies through Prompting}
\label{sec:prompting}
As depicted in Figure~\ref{fig:methods}, this pipeline involves querying the LLMs directly for time series anomalies through a text prompt (as shown below) concatenated with the processed time series window $u^i_{1\dots k} \coloneq \text{prompt} \: \oplus \: (x^i_{1\dots w})$, where $k$ is the total length of the input after concatenation. LLMs will output the next token $u_{k+1}$ sampled from an autoregressive distribution conditioned on the previous tokens $p_\theta(u_{k+1}|u_{1\dots k})$.

Following a series of experiments, as shown in Table~\ref{tab:prompt_examples}, we iterated over trial \#5 and arrived at the following prompt for our study:

\textit{``You are an exceptionally intelligent assistant that detects anomalies in time series data by listing all the anomalies. Below is a sequence, please return the anomalies in that sequence. Do not say anything like `the anomalous indices in the sequence are', just return the numbers. Sequence: \{the input sequence $(x_{1\dots w})$\}.''}

Under this prompt, the LLM generates a list of \textit{values} it delineates as point-wise anomalies. It is noteworthy that the \gpt-3.5-turbo model is capable of directly outputting anomalous indices using the prompt presented in Table~\ref{tab:prompt_examples}, while \mistral{} lacks this ability, as shown in trial \#5. To maintain consistency across our experiments, we conducted experiments on both models using the same prompt mentioned above.

As explained in Section III.A, we adopt the rolling windows method, segmenting the time series into rolling windows before inputting it into the LLMs. For each window, we generate 10 samples from the output probability distribution. For each sample containing values deemed anomalous by LLMs, we collect all indices of the window corresponding to those values. Then, the 10 lists of indices are merged together: if an index appears in at least $\alpha$ percent of the total number of samples, it is considered an anomaly. Finally, the lists of detected anomalies from each window are combined to get the final prediction using a similar criterion: an index is considered an anomaly if it appears in at least $\beta$ percent of the total number of overlapping windows, which are estimated by dividing the window size by the step size. Here, $\alpha$ and $\beta$ are hyperparameters, which can be tuned to improve performance. 

\addtocounter{footnote}{-2} 
\stepcounter{footnote}\footnotetext{\url{https://numpy.org/doc/stable/reference/generated/numpy.convolve.html}}
\stepcounter{footnote}\footnotetext{\url{https://scikit-learn.org/stable/modules/generated/sklearn.ensemble.IsolationForest.html}}

\subsection{\detect: Finding Anomalies through Forecasting}
As depicted in Figure~\ref{fig:how-ml-works}, the first step in a typical ML pipeline involves training an ML model on a collection of time series. From~\cite{gruver2024llmtime}, pretrained LLMs are capable of forecasting time series, allowing us to jump straight to the inference phase.

\noindent\textbf{Pre-processing.}
As detailed in Section~\ref{sec:timeseries-rep}, our first step involves transforming a raw input into a textual representation, and creating samples ready for the LLM from the rolling window sequences $\{\left(x^i_{1\dots w}\right)\}_{i=1}^N$.

\noindent\textbf{Forecasting.}
For each given window $\left(x^i_{1\dots w}\right)$, we aim to predict the next values $(x^i_{w+1\dots w+h})$ where $h$ is the forecast horizon. For ease of notation, the predicted sequence for a window $i$ is noted as $x^i_h$, and the lack of $i$ indication means it is applied for all windows.
This can be achieved through the next token conditional probability distribution noted $p_\theta(x_{h} \,|\, x_{1\dots w} \;\text{and}\; x \in \mathbb{Z}_{\geq 0})$. With this approach, we give the model the input window $\left(x_{1\dots w}\right)$, and sample multiple sequences from the distribution to estimate $\hat{x}_{h} \approx \mathcal{E}^{-1}(\mathcal{G}^{-1}(x_{h}))$. This yields multiple overlapping sequences $\{\left(\hat{x}^i_{1\dots h}\right)\}_{i=1}^N$ at each point in time when $h > 1$. 

\noindent\textbf{Post-processing.}
For each time point $t$, we now have multiple forecasted values $\hat{x}_t$ in different windows when the horizon is larger than $1$, concretely $\{(\hat{x}^{i+h}_t)\}$. We take the median from the collection as the final predicted value for $\hat{x}_t$.
Furthermore, to increase the reliability of the prediction, we take $n$ samples from the distribution for each window $i$. Therefore, each $x_t$ has $n$ samples. To map this back to a univariate time series, we explore the results by taking the mean, median, $5^{th}$-percentile, and $95^{th}$-percentile as values. For the purpose of anomaly detection, an extreme forecast value could indicate a precursor to an anomaly; therefore, acute values can be informative (see Section~\ref{sec:ablation}).
Now, we have reconstructed the time series as $\mathbf{\hat{X}} = (\hat{x}_1, \hat{x}_2, \dots, \hat{x}_T)$.

We next compute the discrepancy between $\mathbf{X}$ and $\mathbf{\hat{X}}$. A large discrepancy indicates the presence of an anomaly. We denote this discrepancy as an error signal $e$ by computing point-wise residuals, given their simplicity and ease of interpretation. We explore the usage of absolute difference suggested by~\citet{hundman2018lstmdt} $e_t = \left|x_t - \hat{x}_t\right|$.
Moreover, we explore how other functions, such as squared difference $e_t = (x_t - \hat{x}_t)^2$ will help reveal the location of anomalies.
More complex functions that capture the difference between two signals, such as dynamic time warping~\cite{muller2007dtw} can be used. However, \citet{geiger2020tadgan} shows that discrepancies found with point-wise errors are sufficient for this purpose. Moreover, we apply an exponentially weighted moving average to reduce the sensitivity of the detection algorithm~\cite{hundman2018lstmdt}. Error values that surpass the threshold are considered anomalous. We use a sliding window approach to compute the threshold to help reveal contextual anomalies that are abnormal compared to the local neighborhood. As such, we assign the window size and step size to $T/3$ and $T/10$ respectively. We set a static threshold for each sliding window as four standard deviations away from the mean. These hyperparameters were chosen based on preliminary empirical results that agree with previous settings in other approaches~\cite{hundman2018lstmdt, geiger2020tadgan, wong2022aer}.

\section{Evaluation}
In this section, we assess our framework and seek to answer the following research questions:
\begin{itemize}
    \item \textbf{RQ1} Are large language models effective anomaly detectors for univariate time series?
    \item \textbf{RQ2} How does the \llmad framework compare to existing approaches?
    \item \textbf{RQ3} What are the success and failure cases and why?
\end{itemize}

\noindent\textbf{Datasets.}
We examined \llmad on 11 datasets with known ground truth anomalies. These datasets were gathered from a wide range of sources, including a satellite telemetry signal corpus from
\textbf{NASA}~\footnote{\url{https://github.com/khundman/telemanom}} that includes two sub-datasets: SMAP and MSL; \textbf{Yahoo S5}~\footnote{\url{https://webscope.sandbox.yahoo.com/catalog.php?datatype=s&did=70}}, which contains four sub-datasets: A1, which is based on real production traffic to Yahoo systems, and three others (A2, A3, and A4) which have been synthetically generated; and \textbf{NAB}~\footnote{\url{https://github.com/numenta/NAB}}, which includes multiple types of time series data from various application domains. We consider five sub-datasets: Art, AWS, AdEx, Traf, and Tweets.
In total, these datasets contain 492 univariate time series and 2,349 anomalies.
The properties of each dataset, including the number of signals and anomalies, the average signal length, and the average length of anomalies, are presented in Table~\ref{tab:data_summary}.
The table makes clear how properties differ between datasets; for instance, the NASA and NAB datasets contain anomalies that are longer than those in Yahoo S5, and the majority of anomalies in Yahoo S5's A3 \& A4 datasets are point anomalies.

\begin{table*}[t]
    \centering
    \caption{Dataset Summary: 492 signals and 2349 anomalies.}
    \label{tab:data_summary}
    \begin{tabular}{lcccc}
    \toprule
  
    Dataset & \# Sub-datasets & \# Signals & \# Anomalies  & Avg. Length \\
                                    \midrule
    \textbf{NASA}        & 2  &  80            & 103   & 8686 $\pm$ 5376   \\
    \textbf{Yahoo S5}       & 4  &  367           & 2152  & 1561 $\pm$ 140 \hspace{1pt}  \\
    \textbf{NAB}         & 5  &  45            & 94    & 6088 $\pm$ 3150  \\
    \midrule
    Total                & 11 &  492           & 2349  \\
  \bottomrule
    \end{tabular}
\end{table*}

\noindent\textbf{Models.}
We used \texttt{Mistral-7B-Instruct-v0.2} for \prompt and \detect. Moreover, we used \texttt{gpt-3.5-turbo-instruct} for \prompt alone. Due to cost constraints, we explored the usage of \gpt{} for \detect on a 5\% sample of all datasets, which produced similar results to using \mistral. 
We compared \llmad to state-of-the-art models in unsupervised time series anomaly detection. This includes a variety of models similar to the ones considered in~\cite{wong2022aer, alnegheimish2022sintel}:
\begin{itemize}
    \item Classic statistical methods including \texttt{ARIMA}, \texttt{Matrix Profiling (MP)}, and a simple \texttt{Moving Average (MAvg)}.
    \item Deep learning models currently considered state-of-the-art, including \texttt{LSTM DT} which is a forecasting-based model, \texttt{LSTM AE}, \texttt{VAE}, and \texttt{TadGAN} which are reconstruction-based models,  and \texttt{AER} which is a hybrid between forecasting and reconstruction. 
    \item \texttt{AnomalyTransformer (AT)}, a transformer architecture model for anomaly detection.
    \item \texttt{MS Azure}, an anomaly detection service.
\end{itemize}
These models use a wide range of underlying detection methods, which increases our anomaly detection coverage overall.

\noindent\textbf{Metrics.}
We utilized anomaly detection-specific metrics for time series data~\cite{tatbul2018precisionrecall, alnegheimish2022sintel}. Namely, we looked at the F1 score, under which both partial and full anomaly detection are considered correct identification.

\noindent\textbf{Hyperparameters.}
For \prompt, GPU capacity means that the maximum input window length of SMAP and MSL is 500 values (for other datasets, it is 200 values). We chose a step size such that, on average, a value was contained in 5 overlapping windows (i.e, 100 steps for SMAP and MSL, and 40 for others).
For \detect, we set the window size to 140 and the step size to 1. With a rolling window strategy of step size 1, it is important to keep the windows as small as possible while still ensuring that they are large enough to make useful predictions, as more context tends to be useful for LLMs. Our preliminary results suggested that a window size of 140 was as performative as a window size of 200, and was better than a window size of 100. We set the horizon to 5.

\noindent\textbf{Computation.}
For \gpt{}, we used \gpt{-3.5-turbo} due to its superior performance on time series data (demonstrated by~\citet{gruver2024llmtime}) and its affordability.
For \mistral, we used the publicly available model hosted by HuggingFace~\footnote{\url{https://huggingface.co/mistralai/Mistral-7B-Instruct-v0.2}} on an Intel i9-7920X 24 CPU core processor and 128GB RAM machine with 2 dedicated NVIDIA Titan RTX 24GB GPUs.
For benchmarking, we use Intel Xeon processor of 10 CPU cores (9 GB RAM per core) and one NVIDIA Volta V100 GPU with 32 GB memory. 

\subsection{Are large language models effective anomaly detectors for univariate time series?}

After running the models on the full datasets, we computed the precision, recall, and F1 scores, shown in Table~\ref{tab:llm-results-summary}. 
Overall, \mistral{} achieved better results than \gpt{} for the \prompt method, with a $2\times$ improvement in F1 score. In addition, \detect performed better overall than \prompt. 
We investigate each approach below. 

\begin{table}[ht]
    \centering
    \caption{Summary of Precision, Recall, and F1 Score}
    \label{tab:llm-results-summary}
    \resizebox{\linewidth}{!}{%
    \begin{tabular}{lccc}
    \toprule 
                        &  Precision    & Recall    & F1 Score \\
    \midrule
    \prompt \mistral    &  0.219 $\pm$ 0.108 & 0.311 $\pm$ 0.213 & 0.223 $\pm$ 0.104 \\
    \prompt \gpt        &  0.162 $\pm$ 0.133 & 0.245 $\pm$ 0.191 & 0.133 $\pm$ 0.076 \\
    \detect             &  \textbf{0.613 $\pm$ 0.184} & \textbf{0.514 $\pm$ 0.211} & \textbf{0.525 $\pm$ 0.167} \\
    \bottomrule
    \end{tabular}}
\end{table}

\noindent\textbf{Ablation Study}
\label{sec:ablation}

\begin{figure}[ht]
    \centering
    \includegraphics[width = \linewidth]{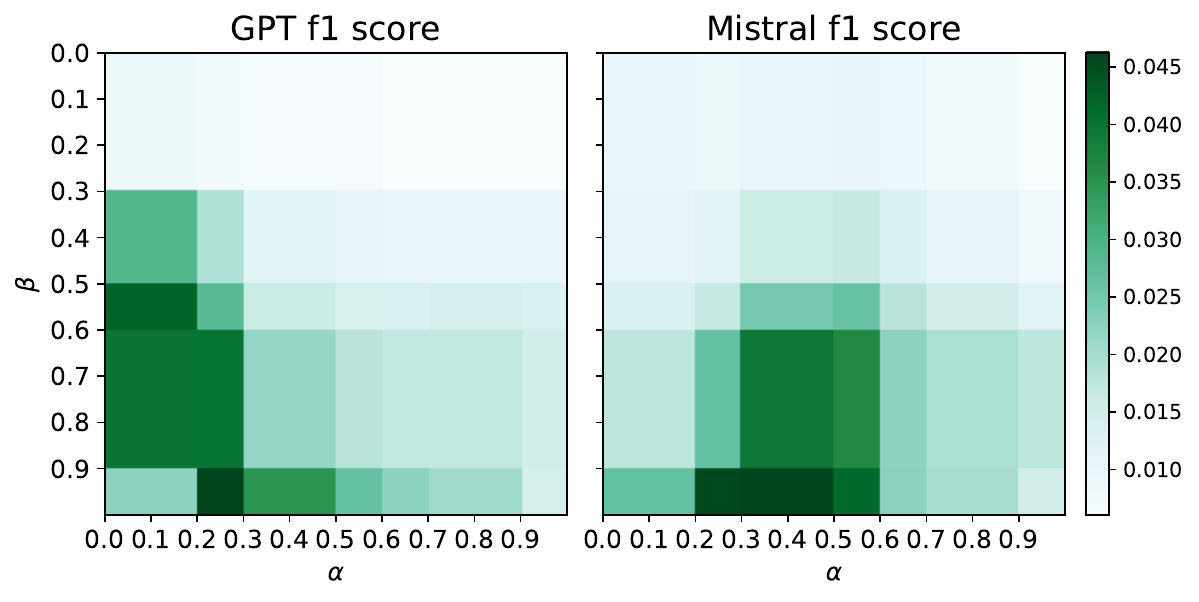}
    \vspace{-0.9cm}
    \caption{Optimizing the choice $\alpha$ and $\beta$ values based on the average F1 scores on all datasets.}
    \label{fig:totalf1}
\end{figure}

\noindent\textbf{\prompt.}
The \prompt approach originally produced an extremely high number of anomalies. We introduced the $\alpha$ and $\beta$ hyperparameters to filter the end result. 
We performed an ablation study to test multiple combinations of $\alpha$ and $\beta$ values on the F1 score. For some windows, the LLMs consistently outputted more than half of the window values as anomalous; thus, we discarded the predicted results of all windows containing all 10 samples, which was more than 50\% of the windows. Fig~\ref{fig:totalf1} shows detection F1 scores from different combinations of $\alpha$ and $\beta$ values. We observed that on all datasets, for \mistral, $\alpha = 0.4$ and $\beta = 0.9$ yielded the best F1 score; for \gpt{}, $\alpha = 0.2$ and $\beta = 0.9$ yielded the best F1 score as shown in Fig~\ref{fig:totalf1}.


\noindent\textbf{\detect}
Since we sampled multiple instances from the probability distribution (namely 10 samples in our experiments), we obtained a possible range of values that could be assigned to each particular point in time. We studied a multiple aggregation function to recreate a one-dimensional signal. Table~\ref{tab:ablation-detector} shows the detection F1 score when the signal is recreated from mean, median, $5^{th}$-percentile, and $95^{th}$-percentile values of the predicted distribution. Moreover, we consider different error scores under smoothing operation and without.

We can see that, on average, squared error with smoothing on the median signal produced the best score, even though it was not the best performer on any individual dataset. The best configuration proved to be different for almost every dataset, with squared error producing the best results on Yahoo S5, and absolute error on NAB. One interesting observation is that the signal reconstructed from the $5^{th}/95^{th}$-percentile showcased higher potential in revealing the location of anomalies.

\begin{table*}[!t]
    \centering
    \vspace{-0.5cm}
    \caption{F1 Score of all variations of \detect}
    \label{tab:ablation-detector}
    \resizebox{\linewidth}{!}{%
    \begin{tabular}{lll*{12}{c}}
    \toprule
    {} & {} &         & \multicolumn{2}{c}{NASA} & \multicolumn{4}{c}{Yahoo S5} & \multicolumn{5}{c}{NAB}  \\
    \cmidrule(lr){3-5}\cmidrule(lr){6-9}\cmidrule(lr){10-14}
    \multicolumn{3}{c}{\textbf{Variation}}  &   MSL &  SMAP &    A1 &    A2 &    A3 &    A4 &   Art &   AWS &  AdEx &  Traf & Tweets & $\mu \pm \sigma$ \\
    \midrule
    \multirow{8}{*}{AE}
    & \multirow{4}{*}{with smoothing}
    &  mean & 0.277 & 0.384 & 0.537 & 0.387 & 0.000 & 0.000 & \textbf{0.400} & 0.329 & 0.640 & 0.444 & 0.586 & 0.362 $\pm$ 0.199 \\
    & &  median & 0.269 & 0.384 & 0.538 & 0.387 & 0.000 & 0.000 & \textbf{0.400} & 0.312 & 0.696 & \textbf{0.480} & 0.593 & 0.369 $\pm$ 0.210 \\
    & &  5\% & 0.294 & 0.400 & 0.542 & 0.387 & 0.000 & 0.000 & \textbf{0.400} & 0.308 & \textbf{0.727} & 0.417 & 0.615 & 0.372 $\pm$ 0.214 \\
    & &  95\% & 0.254 & 0.396 & 0.532 & 0.387 & 0.004 & 0.005 & \textbf{0.400} & 0.289 & 0.696 & 0.348 & 0.655 & 0.361 $\pm$ 0.214 \\
    \cmidrule(lr){2-15}
    & \multirow{4}{*}{w/o smoothing} 
    &  mean & 0.412 & 0.350 & 0.563 & 0.762 & 0.060 & 0.129 & 0.235 & 0.282 & 0.625 & 0.368 & 0.325 & 0.374 $\pm$ 0.200 \\
    & &  median & 0.412 & 0.337 & 0.572 & 0.762 & 0.060 & 0.127 & 0.235 & 0.286 & 0.625 & 0.359 & 0.333 & 0.373 $\pm$ 0.201 \\
    & &  5\% & \textbf{0.429} & 0.353 & 0.564 & 0.759 & 0.040 & 0.123 & 0.235 & 0.284 & 0.621 & 0.400 & 0.350 & 0.378 $\pm$ 0.203 \\
    & &  95\% & 0.406 & 0.337 & 0.608 & 0.765 & 0.114 & 0.167 & 0.235 & 0.288 & 0.600 & 0.387 & 0.342 & 0.386 $\pm$ 0.190 \\
    \midrule
    \multirow{8}{*}{SE}
    & \multirow{4}{*}{with smoothing}
    &  mean & 0.316 & 0.414 & 0.551 & 0.673 & 0.004 & 0.016 & 0.364 & 0.344 & 0.643 & 0.424 & 0.719 & 0.406 $\pm$ 0.228 \\
    & &  median & 0.316 & 0.414 & 0.560 & 0.671 & 0.008 & 0.016 & 0.364 & 0.352 & 0.667 & 0.412 & 0.730 & \textbf{0.410 $\pm$ 0.232} \\
    & &  5\% & 0.333 & 0.400 & 0.552 & 0.662 & 0.000 & 0.014 & 0.364 & \textbf{0.362} & 0.621 & 0.400 & 0.730 & 0.403 $\pm$ 0.227 \\
    & &  95\% & 0.306 & \textbf{0.431} & 0.577 & 0.688 & 0.023 & 0.064 & 0.364 & 0.318 & 0.593 & 0.345 & \textbf{0.762} & 0.406 $\pm$ 0.225 \\
    \cmidrule(lr){2-15}
    & \multirow{4}{*}{w/o smoothing} 
    &  mean & 0.344 & 0.257 & 0.567 & \textbf{0.828} & 0.324 & 0.315 & 0.333 & 0.279 & 0.541 & 0.280 & 0.220 & 0.390 $\pm$ 0.174 \\
    & &  median & 0.344 & 0.247 & 0.583 & 0.824 & 0.323 & 0.315 & 0.333 & 0.281 & 0.541 & 0.259 & 0.220 & 0.388 $\pm$ 0.176 \\
    & &  5\% & 0.358 & 0.241 & 0.563 & \textbf{0.828} & 0.294 & 0.290 & 0.333 & 0.279 & 0.556 & 0.286 & 0.220 & 0.386 $\pm$ 0.178 \\
    & &  95\% & 0.390 & 0.238 & \textbf{0.615} & 0.796 & \textbf{0.376} & \textbf{0.363} & 0.235 & 0.287 & 0.588 & 0.293 & 0.246 & 0.402 $\pm$ 0.176 \\
    \bottomrule
    \end{tabular}}
\end{table*}
\begin{table*}[!t]
\vspace{-0.2cm}
\caption{Benchmark Summary Results depicting F1 Score.}
\label{tab:bench_f1}
\resizebox{\linewidth}{!}{%
\begin{tabular}{lcccccccccccc}
\toprule
& \multicolumn{2}{c}{NASA} & \multicolumn{4}{c}{Yahoo S5} & \multicolumn{5}{c}{NAB}  \\
\cmidrule(r){2-3}\cmidrule(r){4-7}\cmidrule(r){8-12}
\textbf{Pipeline} & MSL & SMAP & A1 & A2 & A3 & A4 & Art & AWS & AdEx & Traf & Tweets & $\mu \pm \sigma$ \\
\midrule
\texttt{AER}      & \textbf{0.587} & \textbf{0.819} & \textbf{0.799} & \textbf{0.987} & \textbf{0.892} & 0.709 & \textbf{0.714} & \textbf{0.741} & 0.690 & \textbf{0.703} & 0.638 & \textbf{0.753 $\pm$ 0.109} \\
\texttt{LSTM DT}  & 0.471 & 0.726 & 0.728 & 0.985 & 0.744 & 0.646 & 0.400 & 0.468 & 0.786 & 0.585 & 0.603 & 0.649 $\pm$ 0.161 \\
\texttt{ARIMA}    & 0.525 & 0.411 & 0.728 & 0.856 & 0.797 & 0.686 & 0.308 & 0.382 & 0.727 & 0.467 & 0.514 & 0.582 $\pm$ 0.176 \\
\texttt{MP}       &	0.474 & 0.423 & 0.507 & 0.897 & 0.793 & \textbf{0.825} & 0.571 & 0.440 & 0.692 & 0.305 & 0.343 & 0.570 $\pm$ 0.193 \\
\texttt{TadGAN}   & 0.560 & 0.605 & 0.578 & 0.817 & 0.416 & 0.340 & 0.500 & 0.623 & 0.818 & 0.452 & 0.554 & 0.569 $\pm$ 0.142 \\
\texttt{LSTM AE}  & 0.545 & 0.662 & 0.595 & 0.867 & 0.466 & 0.239 & 0.667 & \textbf{0.741} & 0.500 & 0.500 & 0.475 & 0.569 $\pm$ 0.158 \\
\texttt{VAE}      & 0.494 & 0.613 & 0.592 & 0.803 & 0.438 & 0.230 & 0.667 & 0.689 & 0.583 & 0.483 & 0.533 & 0.557 $\pm$ 0.143 \\
\texttt{AT}       & 0.400 & 0.266 & 0.571 & 0.565 & 0.760 & 0.576 & 0.414 & 0.430 & 0.500 & 0.371 & 0.287 & 0.467 $\pm$ 0.138 \\
\texttt{MAvg} & 0.171 & 0.092 & 0.713 & 0.356 & 0.647 & 0.615 & 0.222 & 0.408 & \textbf{0.880} & 0.157 & \textbf{0.776} & 0.458 $\pm$ 0.266 \\
\texttt{MS Azure} & 0.051 & 0.019 & 0.280 & 0.653 & 0.702 & 0.344 & 0.056 & 0.112 & 0.163 & 0.117 & 0.176 & 0.243 $\pm$ 0.225 \\
\midrule
\prompt \mistral & 0.160 & 0.154 & 0.194 & 0.235 & 0.338 & 0.336 & 0.370 & 0.268 & 0.000 & 0.135 & 0.257 & 0.223 $\pm$ 0.104 \\
\prompt \gpt{} & 0.049 & 0.110 & 0.143 & 0.078 & 0.157 & 0.195 & 0.154 & 0.194 & 0.133 & 0.133 & 0.197 & 0.133 $\pm$ 0.076 \\
\detect &  0.429 & 0.431 & 0.615 & 0.828 & 0.376 & 0.363 & 0.400 & 0.362 & 0.727 & 0.480 & 0.762 & 0.525 $\pm$ 0.167 \\
\bottomrule
\end{tabular}}
\end{table*}

\begin{figure*}[!t]
\centering
\vspace{-0.25cm}
\includegraphics[width=0.6\linewidth]{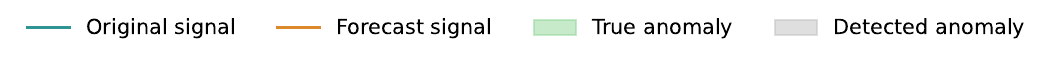}
\begin{minipage}{.45\linewidth}
    \includegraphics[width=\linewidth]{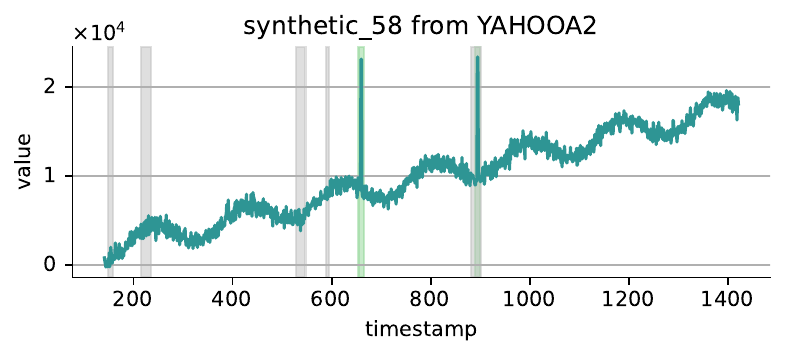}
    \includegraphics[width=\linewidth]{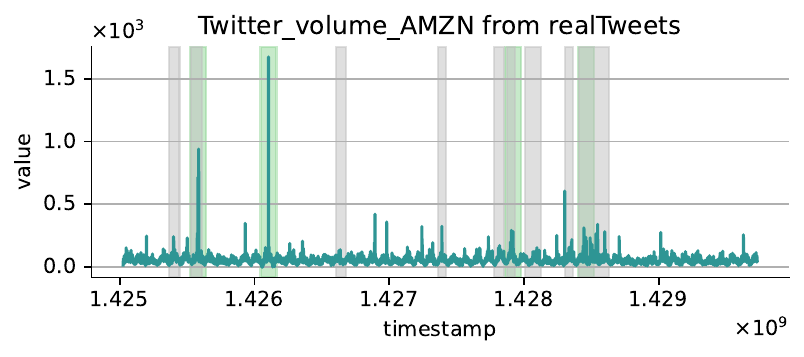}
    \vspace{-0.75cm}
    \caption{Examples of anomalies identified through \prompt. While the model was able to find anomalies, the number of false positives was high, and there were false negatives.}
    \label{fig:prompter-examples}
\end{minipage}%
\qquad
\begin{minipage}{.45\linewidth}
    \includegraphics[width=\linewidth]{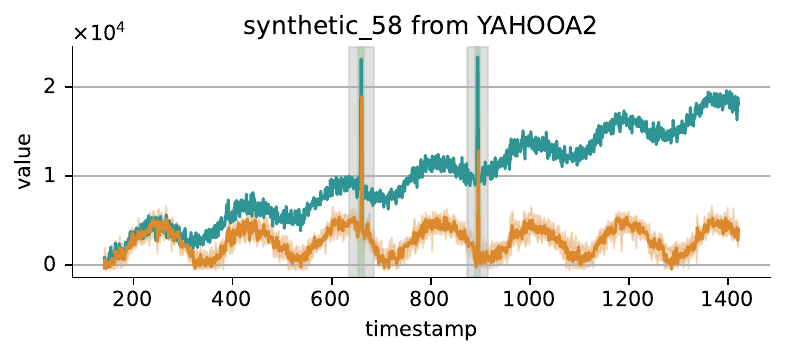}
    \includegraphics[width=\linewidth]{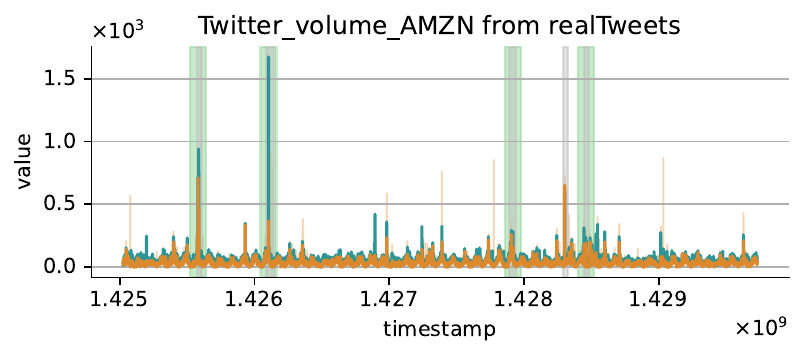}
    \vspace{-0.75cm}
    \caption{Examples of anomalies successfully identified by \detect. Even though the model did not capture the trend present in \texttt{synthetic\_58}, it still managed to find the anomalous intervals.}
    \label{fig:detector-examples}
\end{minipage}%
\end{figure*}

\subsection{How does the \llmad compare to existing approaches?}
Table~\ref{tab:bench_f1} highlights the F1 score obtained for each of the 11 datasets.

\noindent\textbf{LLM-based methods can outperform transformer-based methods by 12.5\%.}
Comparing \detect to \texttt{AT}, which is a transformer-based method, we see \detect outperforms \texttt{AT} in 7 out of the 11 datasets. However, this observation is only relevant to \detect.

\noindent\textbf{LLM-based methods can perform surprisingly well. Our methods achieved an F1 score 14.6\% higher than that of \texttt{MAvg}, and only 10.9\% lower than that of \texttt{ARIMA} which is a reasonable model.}
Moreover, the \detect pipeline alone surpasses \texttt{MAvg} performance in 6 out of 11 datasets, and \texttt{ARIMA} in 4 out of 11 dataset.
We can best see the potential of \detect in the Tweets dataset, where the gap between the LLM's result and the highest result (from \texttt{MAvg}) is minimal, at only $1.8\%$.

\noindent\textbf{\texttt{AER}, the current best deep learning model, is 30\% better than LLM-based approaches. Deep learning methods still perform better than LLM-based methods by 18\% on average.}
Looking more closely at \detect shows that it does not perform as well as deep learning models, achieving a 30\% lower score than that of the highest performing model (\texttt{AER}) and a 5\% lower score than the that of least performative model (\texttt{VAE}). 
Moreover, LLMs do not hold the highest score for any dataset.
It is clear that there is a significant gap here, and an opportunity for improvement.

In the following section, we aim to address where the LLMs succeeded at this problem and where they fell short.

\subsection{What are the success and failure cases and why?}

Figures~\ref{fig:prompter-examples} and~\ref{fig:detector-examples} illustrate example outputs for both the \prompt and \detect methods, respectively. We focus on \mistral{} since it yielded better overall results than \gpt{}, as depicted in Table~\ref{tab:llm-results-summary}. 

\noindent\textbf{Success Cases.}
For both signals shown in Figure~\ref{fig:detector-examples}, \detect correctly identified all the anomalies, with only one false alarm in the second signal. 
The \prompt approach is able to detect outliers and locally extreme values a lot more effectively than anomalies that are buried within the signal. For example, if the window is ``244 , 309 , 2 , 462'', both \gpt{} and \mistral{} point to ``2'' as the observed anomaly. However, this problem becomes more ambiguous when the context is larger. As seen in Figure~\ref{fig:prompter-examples} on Twitter\_volume\_AMZN signal, \prompt identified some locally extreme values as anomalous. Oddly enough, it also missed some anomalies that were global outliers.

\noindent\textbf{Failure Cases.}
Even though \detect correctly identified all anomalies in the example shown in Figure~\ref{fig:detector-examples}, the forecast itself struggled to capture the non-stationary aspects of the signal, particularly its trend. This is due to the sensitivity of LLMs to context length. A window size larger than 140 is needed to capture this property. While it did not impact the detection in this case, this may explain failure cases in other signals.

\prompt raised a large number of false alarms, with an average precision of 0.219.
Using the filtering method described in Section~\ref{sec:prompting} does not eliminate false positives. An alternative strategy could be to use log probabilities as a measure of confidence for filtering. We recommend exploring this avenue in future work.







\section{Discussion}

\noindent\textbf{Prompting Challenges.}
Over a three-month experimental period, various prompts were employed, as laid out in Table~\ref{tab:prompt_examples}. It is evident that both \gpt{} and \mistral{} fail to produce the desired responses unless a chat template is applied that attributes roles to the user and the system. Furthermore, to ensure the exclusivity of numerical values in the generated responses, in addition to specifying in the prompt to ``just return numbers,'' we adjusted the likelihood of non-numerical tokens appearing in the output generated by the LLMs.

Under the `find indices' prompt, \gpt{} may generate lists of indices; however, these indices frequently surpass the sequence length. Conversely, \mistral{} yields values instead of indices when utilizing the same prompt. Therefore,  for our experiment, we altered the prompt to include "find values" rather than "find indices."

Unlike \mistral{}, \gpt{} outputs a ``repetitive prompt'' error when presented with a series of identical values within a window. This happened particularly for NASA datasets (there are 23 signals in SMAP and 13 in MSL with this error, affecting up to 85\% of the windows). In this experiment, we deemed such windows as having no detectable anomalies, obtaining a true positive of zero.


\noindent\textbf{Addressing Memorization.}
Large language models are trained on a vast amount of data. The training data for most models -- for instance, those provided by OpenAI -- is completely unknown to the general public, which makes evaluating these models a nuanced problem. Given that large language models, especially \gpt{} models~\cite{chang2023speakmemory}, are notorious for memorizing training data~\cite{biderman2023emergentmemorization}, how do we ensure that there was no data and label leakage for the benchmark datasets used? We posit that our transformation of the time series data into its string representation is unique, essentially making the input time series different from its original form and reducing the chances of blatant memorization. Moreover, unlike with the forecasting task, the task of anomaly detection is not inherent to the training convention used, which is next token prediction.

\begin{figure}[ht]
    \centering
    \includegraphics[width=0.48\linewidth]{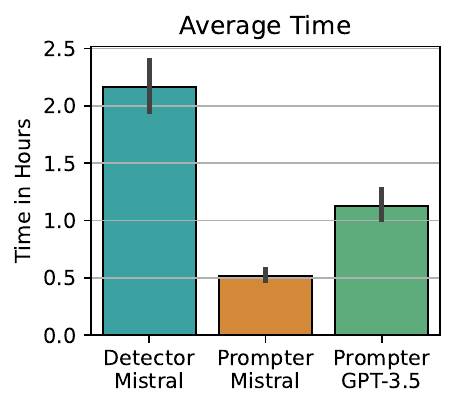}
    \includegraphics[width=0.48\linewidth]{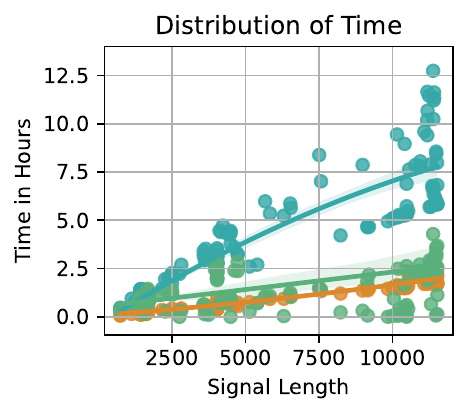}
    \caption{Recorded time for \prompt and \detect. (left) On average, \detect takes the longest to infer, almost double the time of \prompt. (right) Distribution of signal length and execution time.}
    \label{fig:infer-time}
\end{figure}

\noindent\textbf{Practicality of Usage.}
The appeal of using LLMs for this task lies in their ability to be used in zero-shot, without necessitating any fine-tuning. However, this property is bottlenecked by the latency time. Figure~\ref{fig:infer-time} illustrates the average time it takes to use LLMs for each of the suggested approaches. It is unreasonable to wait half an hour to two hours for the model to produce a response, especially when deep learning models take less than an hour to train.
Since these experiments were run in an offline setting, we can expect that real-time deployment would be more sensible, especially since the context size is much smaller, eliminating the need for rolling windows. In other cases, a single window would be sufficient, which takes approximately 5 seconds to infer depending on the window size. 

In total, running \prompt experiments using the \texttt{gpt-3.5-turbo-instruct} version has cost us approximately \$834.11 -- an average of \$1.69 per signal. For \detect, we ran a small-scale experiment where we sampled 22 signals of the data, roughly 5\%. The total reached \$95.08 -- an average of \$4.3 per signal -- making \detect a more expensive method than \prompt.


\vspace{-0.2cm}
\section{Conclusion}
In this paper, we present large language models with a new and challenging task: detecting anomalies in time series data with no prior learning. To this end, we propose two methods, \prompt and \detect, covering the prompting and forecasting paradigms respectively.
We demonstrate that LLMs can find anomalies through the forecasting paradigm (\detect method) more accurately than they can through the \prompt method.
We present \llmad, a framework for converting signals into text, enabling LLMs to work with time series data. A major weakness of LLMs is their limited context window size. In \llmad, we rely on rolling windows to chop the time series up into smaller segments. This is both inefficient and costly. Moreover, it seriously challenges the possibility of  expanding the framework to work with multivariate time series. As LLMs rapidly advance, more models can handle larger context sizes; however, they do not yet have the capacity to handle time series data without segmentation.
Even with their limited capability, LLMs are able to beat a well-known transformer method, and to approach the performance of classical methods when tested against 492 signals. They fall short of deep learning models by a factor of $\times1.2$.
In anomaly detection, post-processing strategies are critical for revealing the locations of anomalies. In future work, we plan to investigate post-processing functionalities that can help \prompt filter false alarms. Similarly, we plan to conduct a thorough exploration of error functions that bring out anomalies for \detect.

\section*{Acknowledgments}
Our research is supported by SES S.A., Iberdrola and ScottishPower Renewables, and Hyundai Motor Company.

\section*{Change Log}
Updated the acknowledgment section. Removed the previous version which had acknowledgements to the folks who had helped create the IJCAI template.

\bibliographystyle{named}
\bibliography{ref}


\end{document}